Zhengyang Duan[1, †], Hang Chen[1, †], Xing Lin[1, 2, 3, 4, *]

# Optical multi-task learning using multi-wavelength diffractive deep neural networks

[1] Department of Electronic Engineering, Tsinghua University, Beijing 100084, China

[2] Beijing National Research Center for Information Science and Technology, Tsinghua University, Beijing 100084, China

[3] Institute for Brain and Cognitive Sciences, Tsinghua University, Beijing 100084, China

[4] Beijing Laboratory of Brain and Cognitive Intelligence, Beijing Municipal Education Commission, Beijing 100084, China

† These authors contributed equally to this work.

* Corresponding author. Email: lin-x@tsinghua.edu.cn (X.L.)

**Abstract:** Photonic neural networks are brain-inspired information processing technology using photons instead of electrons to perform artificial intelligence (AI) tasks. However, existing architectures are designed for a single task but fail to multiplex different tasks in parallel within a single monolithic system due to the task competition that deteriorates the model performance. This paper proposes a novel optical multi-task learning system by designing multi-wavelength diffractive deep neural networks (D$^2$NNs) with the joint optimization method. By encoding multi-task inputs into multi-wavelength channels, the system can increase the computing throughput and significantly alleviate the competition to perform multiple tasks in parallel with high accuracy. We design the two-task and four-task D$^2$NNs with two and four spectral channels, respectively, for classifying different inputs from MNIST, FMNIST, KMNIST, and EMNIST databases. The numerical evaluations demonstrate that, under the same network size, multi-wavelength D$^2$NNs achieve significantly higher classification accuracies for multi-task learning than single-wavelength D$^2$NNs. Furthermore, by increasing the network size, the multi-wavelength D$^2$NNs for simultaneously performing multiple tasks achieve comparable classification accuracies with respect to the individual training of multiple single-wavelength D$^2$NNs to perform tasks separately. Our work paves the way for developing the wavelength-division multiplexing technology to achieve high-throughput neuromorphic photonic computing and more general AI systems to perform multiple tasks in parallel.

## 1 Introduction

Photonic computing utilizes photons instead of electrons for computation, which possesses the inherent advantages of light-speed processing, low-power consumption, and high-throughput capability [1-5]. Photonic neural networks (PNNs) [6-8], which implement the artificial neural network model based on photonic computing, can achieve leapfrog improvement in computing speed and energy efficiency. Therefore, it's considered one of the most promising solutions to support the sustainable development of artificial intelligence (AI) in the post-Moore era [9-10]. Among different PNN architectures, diffractive deep neural networks (D$^2$NN) [11,12] can achieve large-scale neural information processing and have attracted a vast amount of interest. D$^2$NN consists of layers of diffractive neurons and their optical interconnections based on the diffraction of light, which can be trained with deep learning optimization methods to fit desired mapping functions between input optical fields and output detector measurements.

In the previous studies of D$^2$NN architecture [11-15], monochromatic plane waves are used to encode the input data and propagate through modulation layers, performing a specific task. Inspired by biological intelligence [16], the multiplexing of different tasks in a single D$^2$NN system is of great importance in improving its generalization and expanding its applications for different scenarios. However, performing multiple AI tasks in parallel with a monolithic D$^2$NN system remains challenging. One of the major obstacles is the competition among tasks during the training, which leads to catastrophic forgetting [17,18]. Catastrophic forgetting occurs when systems are trained on multi-tasks, which leads to a tendency for knowledge of previously learned tasks to be abruptly lost while learning new tasks, resulting in the deterioration of performance on every single task. Previous solutions [19,20] require the mechanical movement of optical elements to switch between tasks one at a time or require the design of multiple different D$^2$NN systems, one for each task, significantly increasing the hardware complexity.

Here, we propose an optical multi-task learning monolithic system design that can simultaneously perform multiple classification tasks on different databases without mechanical movement by developing multi-wavelength D$^2$NNs.

Different from previous broadband D²NNs [21,22], the wavelength dimension is exploited in this work to improve the computing throughput, which encodes different inputs into different wavelength channels and performs photonic computing in both spatial and spectral dimensions. We demonstrate that the multi-wavelength D²NNs allow for high-parallel processing of multiple inputs and significantly alleviate the competition among different tasks to preserve the high performance of each task. In this design, the multi-wavelength D²NN has $N$ ($N \geq 2$) different parallel wavelength channels, encoding $N$ different inputs in parallel, and the detection area of each category is segmented into $N$ parts, where each part represents the category of input encoded at the corresponding wavelength channel. We use the multi-wavelength joint optimization method with the loss functions of softmax cross-entropy and energy efficiency constraint to train the D²NN. We first verify the high-parallel characteristic of a three-wavelength D²NN by classifying three different inputs in parallel based on the MNIST database, where the accuracy at each wavelength is comparable to training three single-wavelength D²NN with sequential inputs. To perform multiple tasks in parallel, we encode the inputs from different databases into different wavelength channels. We utilize the two-wavelength and four-wavelength D²NNs for performing the two-task and four-task classifications, respectively, based on the databases of MNIST, FMNIST, KMNIST, and EMNIST. With the increase in task numbers, the multi-wavelength D²NNs achieve significantly higher classification accuracy than the single-wavelength D²NNs and maintain the model accuracy for each task with larger network sizes, demonstrating the great advantages of multi-wavelength D²NNs in realizing optical multi-task learning.

## 2 Methods

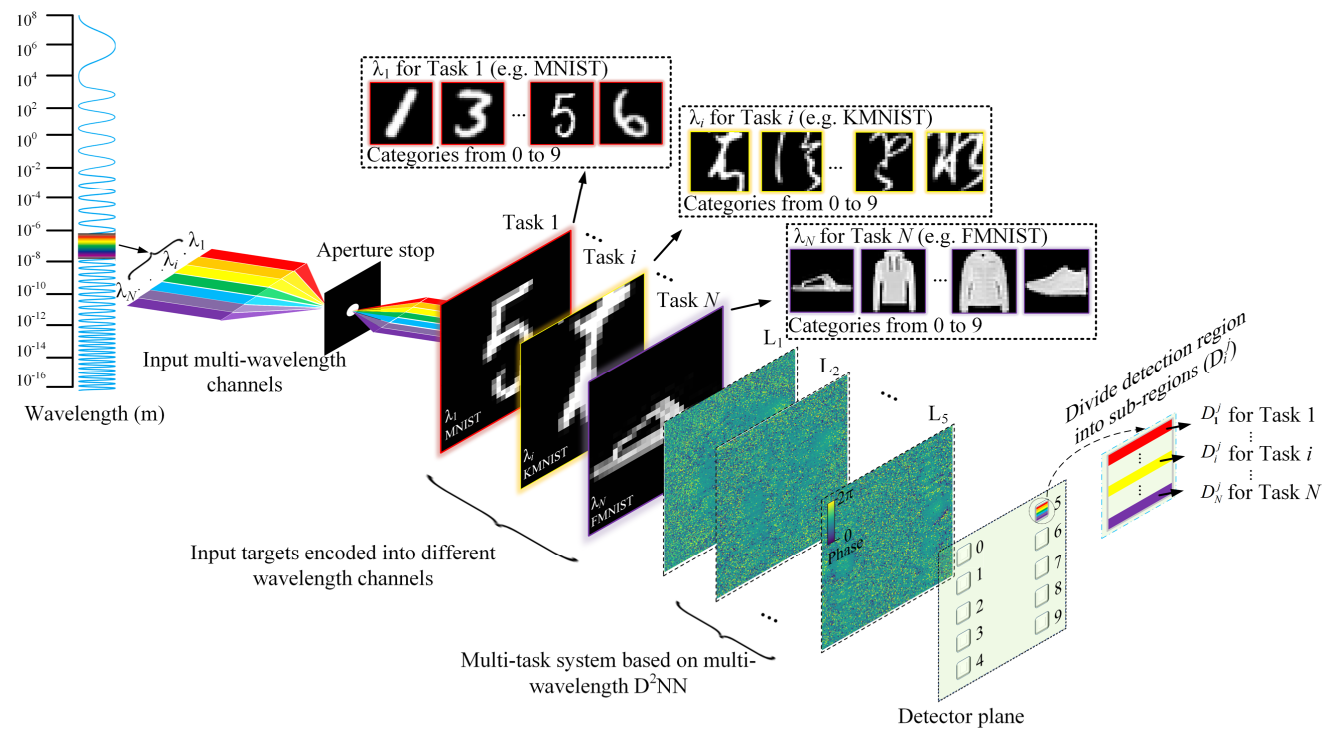

**Fig. 1.** The architecture of multi-wavelength D²NNs for optical multi-task learning. The incident light comprises multiple wavelength channels, where each detection region is correspondingly segmented into multiple sub-regions. Each sub-region represents an input category type for each task.

As shown in Fig. 1, the proposed optical multi-task learning system achieves multiple tasks in parallel by designing a multi-wavelength D²NN. Each wavelength channel ($\lambda_i, i = 1, \ldots, N$) encodes the input targets of each task $i$. By using the approximation theory of multi-wavelength optical systems [23-26], the transformation of multi-wavelength optical fields can be recognized as a combination of independent transformation of coherent optical fields at each wavelength, following the principle of superposition of optical intensities. The input optical fields $U_{\lambda_i}$ at the wavelength of $\lambda_i$, encoding the input targets of task $i$, are transformed by the D²NN before the detection. We consider the linear D²NN in this work with the complex transform function of $M(\Phi)$, where $\Phi$ represents the phase modulation coefficients of the diffractive elements at multiple phase-only diffractive layers. The detailed derivation of complex transform function can be found in [11,12,27]. We assume that the phase modulation coefficients of each diffractive layer are the same under different wavelengths with the multi-wavelength diffractive optical element (DOE) design (see Discussion section).

Therefore, the output optical fields at the $i$-th wavelength $\lambda_i$ can be formulated as: $\boldsymbol{U}_i' = \boldsymbol{M}_i(\boldsymbol{\Phi})\boldsymbol{U}_i$, and the detector measures the intensity distribution of output optical fields that can be formulated as: $\boldsymbol{I}_i = |\boldsymbol{U}_i'|^2 = |\boldsymbol{M}_i(\boldsymbol{\Phi})\boldsymbol{U}_i|^2$. For the multi-wavelength D$^2$NN, the total intensity distribution of different wavelengths can be formulated as the superposition of detected intensity distribution at each wavelength: $\boldsymbol{I} = \sum_i \boldsymbol{I}_i = \sum_i |\boldsymbol{M}_i(\boldsymbol{\Phi})\boldsymbol{U}_i|^2$.

We develop the joint optimization method to train multi-wavelength D$^2$NNs for performing optical multi-task learning. Each category detection area at the output plane is divided into multiple sub-regions $\{D_i^j\}$, where $i = 1, \dots, N$ denotes the index of the task encoded at $i$-th wavelength $\lambda_i$; and $j = 1, \dots, M$ denotes the index of detection areas, representing the index of categories. We calculate the average intensity of the $i$-th sub-region among the $M$ category detection areas, i.e., $\boldsymbol{P}_i = \{avg(\boldsymbol{I}(D_i^j)), j = 1, \dots, M\}$, where $\boldsymbol{I}(D_i^j) = \sum_{\lambda_i} \boldsymbol{I}_{\lambda_i}(D_i^j)$ is for the broadband wavelength detection without using spectral filters on the detector. Besides, the wavelength selective filter can be applied to each sub-region to eliminate the crosstalk during intensity detections among wavelength channels for further improving the task performance. In this case, each sub-region for each task only detects the optical signals at the corresponding wavelength channel, with which $\boldsymbol{I}(D_i^j) = \boldsymbol{I}_i(D_i^j)$. The category type of inputs for the $i$-th task is determined by finding the sub-region of the detection area with maximum average intensity, i.e., the index of the maximum value in the vector $\boldsymbol{P}_i$. Besides, we further include the constraint to maximize the energy transmission efficiency of multi-wavelength D$^2$NNs by minimizing the optical energy outside the category detection areas. Therefore, the joint optimization problem for multi-wavelength D$^2$NNs training can be formulated as:

$$\min_{\boldsymbol{\Phi}} \sum_i (L(\boldsymbol{P}_i, \boldsymbol{G}_i) + MSE(\boldsymbol{I} - \sum_j \boldsymbol{I}(D_i^j))), \tag{1}$$

where $L(\boldsymbol{P}_i, \boldsymbol{G}_i)$ represents the softmax cross-entropy loss function of $i$-th task at the wavelength of $\lambda_i$ between the detection $\boldsymbol{P}_i$ and ground truth label $\boldsymbol{G}_i$; $\boldsymbol{G}_i$ is a one-hot vector with a length of $M$; $MSE(\boldsymbol{I} - \sum_j \boldsymbol{I}(D_i^j))$ represents the total energy of optical intensity outside the sub-regions of category detection areas evaluated with the mean square error.

In the process of training, different wavelength channels share the same phase modulation coefficients at each diffractive layer that are iteratively updated to perform multi-task functions by solving the joint optimization problem in Eq. (1). We use the stochastic gradient descent approach to train the multi-wavelength D$^2$NNs. The input targets of the training datasets of different tasks are encoded into the amplitude of optical fields at different wavelengths to feed into the network input layer. The residual error of network outputs with respect to ground truth labels and the total optical energy outside the category detection areas are calculated according to Eq. (1), which are used to perform the error back-propagation to optimize the network structure and the phase modulation coefficients of optical diffractive elements.

# 3 Results

## 3.1 Multi-wavelength D$^2$NNs for high-parallel classification

We first verify the application of multi-wavelength D$^2$NNs for high-parallel classification that can simultaneously classify multiple inputs in performing a single task. To demonstrate, we use the PyTorch deep learning framework to build a three-wavelength D$^2$NN architecture with five phase-only diffractive modulation layers, which is applied for classifying the MNIST database that can recognize three handwritten digits at each instant of time. We consider the visible wavelength ranging from 400 nm to 700nm, where the input light source was set as the combination of three wavelengths of 400 nm, 550 nm, and 700 nm, encoding three handwritten digits, respectively. Therefore, each category detection area at the output plane is correspondingly segmented into three sub-regions. The Adam optimizer is used for network training to optimize the phase modulation coefficients $\boldsymbol{\Phi}$ of the optical diffractive elements. Each optical diffractive element size was set to 4 $\mu m \times 4$ $\mu m$. We first evaluate the performance of multi-wavelength D$^2$NNs by setting the modulation element number at each network layer to 200 × 200, corresponding to the network layer size of 0.8 $mm$ × 0.8 $mm$ (see Fig. 2b). We further evaluate and compare the network performance under different modulation element numbers at each layer, i.e., K × K, K=200, 400, 600, 800 (see Fig. 2d). The layer number was set to 5, and the distance between successive layers was optimized according to the maximum half-cone diffraction angle theory [11, 27]. With a training batch size of 32, the initial learning rate is set to 0.01 and is reduced by half, i.e., multiplied by 0.5, after every epoch during the training. The network training converges after five epochs to achieve the desired mapping function for the multi-channel inputs and output. The network was trained with 60,000 handwritten digits and blindly tested with 10,000 handwritten digits. For the network layer with a modulation element number of K × K, each digit with a pixel number of 28 × 28 is resized to K/2 × K/2 and padded to K × K.

The numerical evaluation results are shown in Fig. 2, where the performance of multi-wavelength D²NNs is validated with and without wavelength selective filters on each category detection area. Fig. 2(a-c) shows an exemplar result of simultaneously classifying three handwritten input digits, i.e., "7", "2", and "5", encoded in the wavelengths of 700 nm, 550 nm, and 400 nm, respectively, under the modulation element numbers of 200 × 200 at each layer. The classification result of each wavelength channel is determined by finding the maximum average intensity value among the corresponding sub-regions of category detection areas, indicated with three white arrows for three input digits, as shown in Fig. 2(b-c, left). The energy distributions of the classification results of three inputs at different wavelength channels in Fig. 2(b-c) show that the proposed system could prominently identify the sub-region with maximum average intensity for the correct categorization. Due to the use of wavelength selective filters to eliminate the wavelength crosstalk during the detection, the classification accuracies of multi-wavelength D²NNs with wavelength selective filters are 95.9%, 96.4%, and 96.9% for the wavelengths of 700 nm, 550 nm, and 400 nm, respectively, which are slightly higher than the classification accuracies of broadband wavelength detection without wavelength selective filters, i.e., 95.0%, 95.7%, and 96.4%, respectively. For both network settings, the classification accuracies of multi-wavelength D²NNs further improve at each wavelength with the increase of the modulation element numbers at each network layer, as shown in Fig. 2(d). Under the modulation element number of 800 × 800 at each layer, the classification accuracies of multi-wavelength D²NNs with wavelength selective filters, achieving 98.2%, 98.1%, 98.1% for the wavelengths of 700 nm, 550 nm, and 400 nm, respectively, are comparable to training three single-wavelength D²NNs with the serial inputs, i.e., sequential input of digits. The results verify that multi-wavelength D²NN can significantly increase the parallel computing capability. Using multi-wavelength D²NNs for multi-task learning by encoding different tasks into different channels enables different machine learning tasks to be implemented in parallel within a single system.

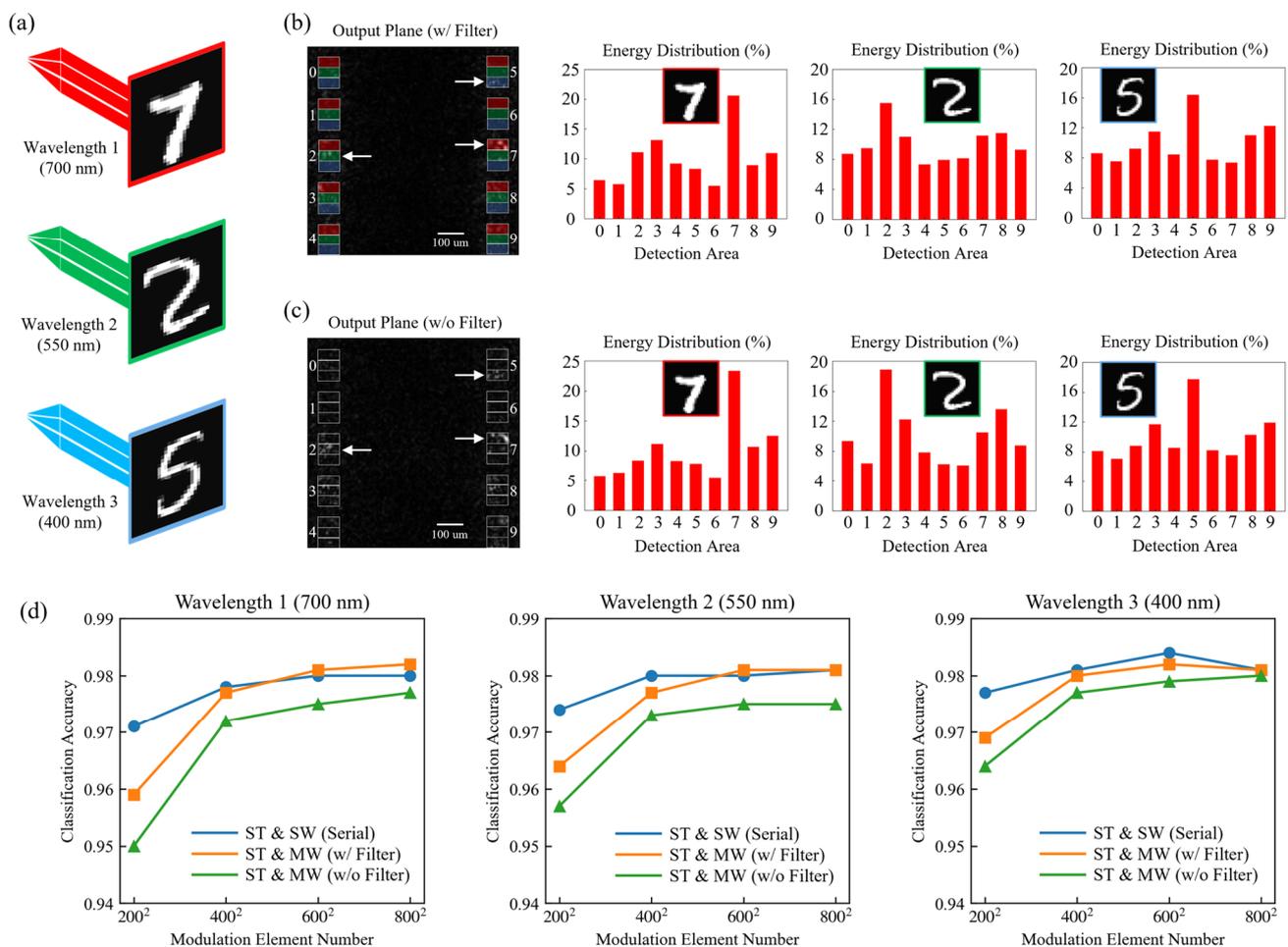

**Fig. 2.** Multi-wavelength D²NNs for high-parallel classification. (a-c) The exemplar results of simultaneously classifying three handwritten digits encoded in three wavelength channels. The classification results are evaluated with and without the wavelength selective filters on each category detection area. (d) The comparison of performances between the three-wavelength D²NNs for high-parallel classification and training three single-wavelength D²NNs with serial inputs under different network layer sizes. The results demonstrate the capability of multi-wavelength D²NNs for high-parallel classification. ST & SW: single task using single wavelength with serial inputs; ST & MW: single task using multiple wavelengths with parallel inputs.

## 3.2 Optical Multi-task Learning using multi-wavelength D²NNs

To demonstrate the capability of multi-wavelength D²NNs for optical multi-task learning, we first construct a two-task classifier for classifying both the MNIST database (task I) and the fashion-MNIST (FMNIST) database (task II). Both databases include 60,000 training samples and 10,000 testing samples with ten category numbers. Therefore, the two-wavelength D²NN was constructed by dividing each of ten detector areas into two sub-regions, where the upper and lower regions represent the classification results of tasks I and II, respectively, as shown in Fig. 3. The handwritten digits of task I are encoded in the wavelength of 700 nm and the fashion products of Task II are encoded in the wavelength of 400 nm. With other network settings the same in Fig. 2, we first set the two-wavelength D²NN to have five diffractive layers, each layer with a modulation element number of 200 × 200, without the wavelength selective filters on the detector. Fig. 3(a) shows an exemplar result for simultaneously classifying a handwritten digit "7" with the category number of 7 from the MNIST database and a fashion product "pullover" with the category number of "2" from the FMNIST database. The energy distributions of the classification results of two tasks in Fig. 3(a, right) show that the multi-wavelength D²NN can prominently identify the sub-region with maximum average intensity for the correct categorization. The maximum average intensity outputs of task I and task II were focused on the upper sub-regions of the No.7 detector area and the lower sub-regions of the No.2 detector area, respectively, which were marked by the white arrow in Fig. 3(a).

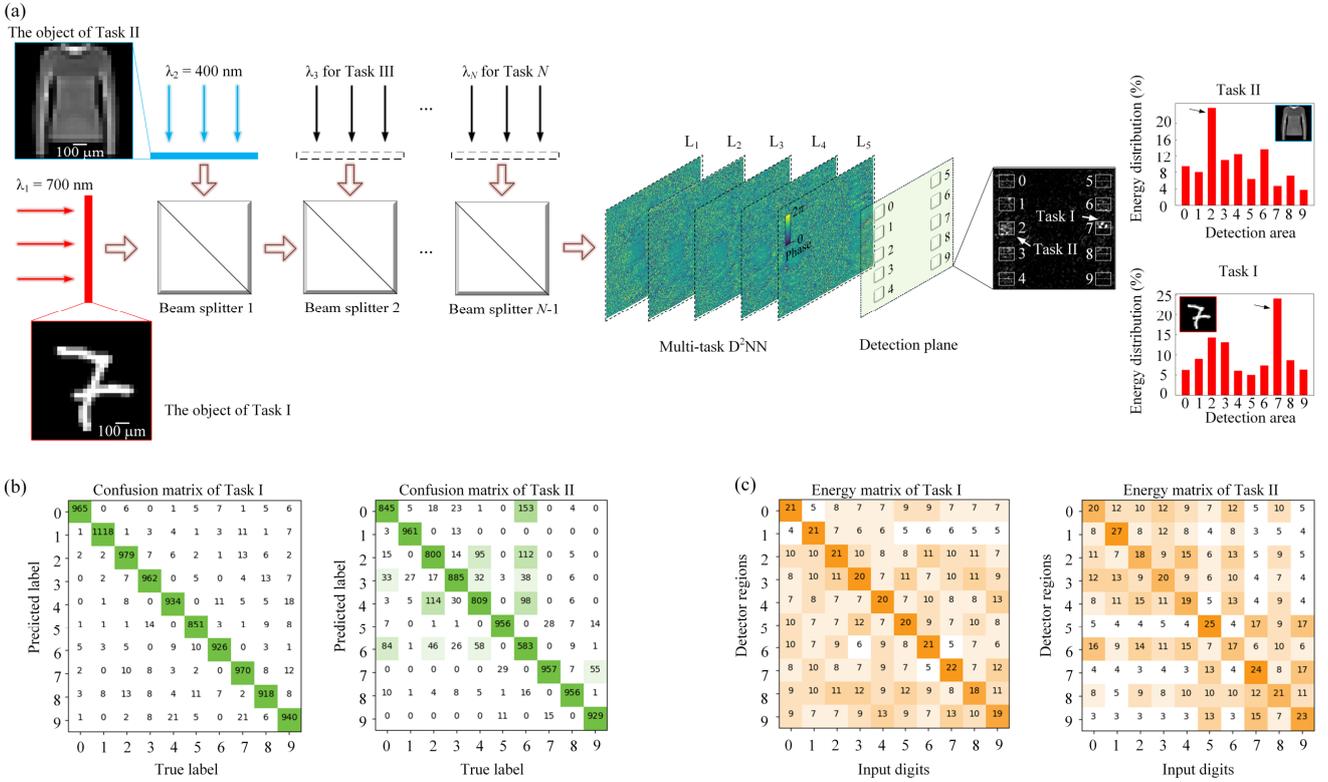

**Fig. 3.** Multi-wavelength D²NNs working under two wavelengths for classifying both the MNIST and FMNIST databases. (a) Images from the MNIST and FMNIST datasets are encoded in wavelengths of 700 nm and 400 nm, respectively. The categories of two inputs are determined by finding the corresponding sub-regions with maximum average intensity as indicated by the white arrow. Energy distributions of the classification results of two tasks demonstrated the success of the proposed approach for classifying two inputs. (b, c) Confusion matrices and energy matrices (percentage) of two tasks, corresponding to the classification accuracies of 95.6% and 86.8% for Task I and Task II, respectively.

The blind testing of the trained two-wavelength D²NN model on the test datasets of MNIST and FMNIST achieves classification accuracies of 95.6% and 86.8%,, respectively. The corresponding confusion matrices and energy distribution matrices, statistically summarizing the classification results of all samples and energy distribution percentages of two tasks, are shown in Fig. 3(b) and (c), respectively. The average energy percentages of correct categories are 20.8% and 21.8% for two tasks, respectively. We further compare the performance of two-wavelength D²NNs for performing two tasks in parallel with respect to the single-wavelength D²NNs for performing two tasks in parallel by overlapping to multiplex two images from two datasets, respectively, as the network input, as shown in Table 1. The classification accuracies of single-wavelength D²NNs for performing two tasks in parallel are 92.4% and 83.1%, respectively, which is much lower than the two-wavelength D$^2$NNs. We also train two single-wavelength D$^2$NNs for individually performing each of the two tasks, where the classification accuracies are 97.1% and 87.5% for tasks I and II, respectively. To improve the performance of two-wavelength D$^2$NN, we can increase the modulation element numbers at each layer, which achieves the classification accuracies of 97.5% and 88.0% for two tasks with $400 \times 400$ modulation element numbers per layer. The performance can be further improved by using the wavelength selective filters on the category detection regions, which achieves the classification accuracies of 95.9% and 87.0% for two tasks with $200 \times 200$ modulation element numbers per each layer and 97.6% and 88.9% for two tasks with $400 \times 400$ modulation element numbers per each layer, showing comparable and even higher accuracy than individually training two single-wavelength D$^2$NNs to perform two tasks separately. The results are summarized in Table 1, which verifies that the designed two-wavelength D$^2$NN with a joint training approach can successfully classify the targets from two tasks in parallel without any mechanical adjusting of diffractive layers for two tasks.

| | Diffractive neural network size | Accuracy | |
| --- | --- | --- | --- |
| | | MNIST(Task I) | FMNIST(Task II) |
| Single-task, Single-wave | 200×200×5 | 97.1% | 87.5% |
| Multi-task, Single-wave | 200×200×5 | 92.4% | 83.1% |
| Multi-task, Multi-wave (w/o Filter) | 200×200×5 | 95.6% | 86.8% |
| Multi-task, Multi-wave (w/ Filter) | 200×200×5 | 95.9% | 87.0% |
| Multi-task, Multi-wave (w/o Filter) | 400×400×5 | 97.5% | 88.0% |
| Multi-task, Multi-wave (w/ Filter) | 400×400×5 | 97.6% | 88.9% |

**Table 1.** Performance comparisons between the multi-wavelength and single-wavelength D$^2$NNs for two-task classifications.

To demonstrate the capability of multi-wavelength D$^2$NNs for multi-task learning with more number of tasks, we constructed a four-wavelength D$^2$NN for four-task classification that can simultaneously classify four targets from the databases of MNIST (task I), FMNIST (task II), Kuzushiji-MNIST (KMNIST, task III), and Extended-MNIST (EMNIST, task IV), respectively. The KMNIST comprises images of ancient Japanese scripts with the same dataset size and category numbers as the MNIST and FMNIST databases. We randomly selected ten categories of handwritten letters from the EMNIST database and kept the same dataset size as the other three tasks, i.e., 60,000 training samples and 10,000 testing samples. The databases of four tasks, from the task I to IV, are encoded in the wavelengths of 700 nm, 600 nm, 500 nm, and 400 nm, respectively. In this numerical experiment, the four-wavelength D$^2$NNs are designed without using wavelength selective filters that have lower hardware complexity. With other network settings the same as Figs. 2 and 3, we evaluated the classification accuracies of four-wavelength D$^2$NNs in performing four tasks in parallel under different network sizes and compared the classification accuracies with the single-wavelength D$^2$NNs, as shown in Fig. 4. For the four-wavelength D$^2$NN with five layers and the modulation element number of 200×200 at each layer, the classification accuracies of four tasks, from the task I to IV, are 92.8%, 83.0%, 81.0%, and 90.4% respectively, which are significantly higher than the single-wavelength D$^2$NN of 64.6%, 68.7%, 52.5%, and 55.3%, under the same network size. The four-wavelength D$^2$NNs for four-task classification consistently achieved much higher accuracies than the single-wavelength D$^2$NNs with varying network sizes. As the number of tasks increases from two to four, the proposed multi-wavelength D$^2$NN shows more advantages in realizing optical multi-task learning.

We further evaluated and compared the performance of the proposed four-wavelength D$^2$NNs with respect to the individual training of four single-wavelength D$^2$NNs to perform four tasks separately (see Fig. 4) under different network sizes. Fig. 4(a) increases the network size by increasing the layer numbers from 1 to 8 with the same element number of 200×200 at each modulation layer. Fig. 4(b) increases the network size by increasing the element number at each modulation layer with the same layer number of 5. Increasing the neural network size of multi-wavelength D$^2$NNs for optical multi-task learning can significantly improve its inference capability until the performance reaches a state of saturation. The performance of four-wavelength D$^2$NNs continues to improve with the increase of network size and approaches to the performance of training four single-wavelength D$^2$NNs. The classification accuracies of task I to task IV are 96.5%, 85.6%, 88.6%, and 93.8%, respectively, with the modulation layer number of five and the element number of 800×800 at each layer, which shows comparable performance with respect to the training of four single-wavelength D$^2$NNs with the same network size. The results demonstrate the effectiveness of the proposed approach for multi-task learning with a monolithic optical system and achieve much lower hardware complexity. The encoding of multi-tasks into multi-wavelength channels alleviates the competition among different tasks and minimizes the performance reduction of each task.

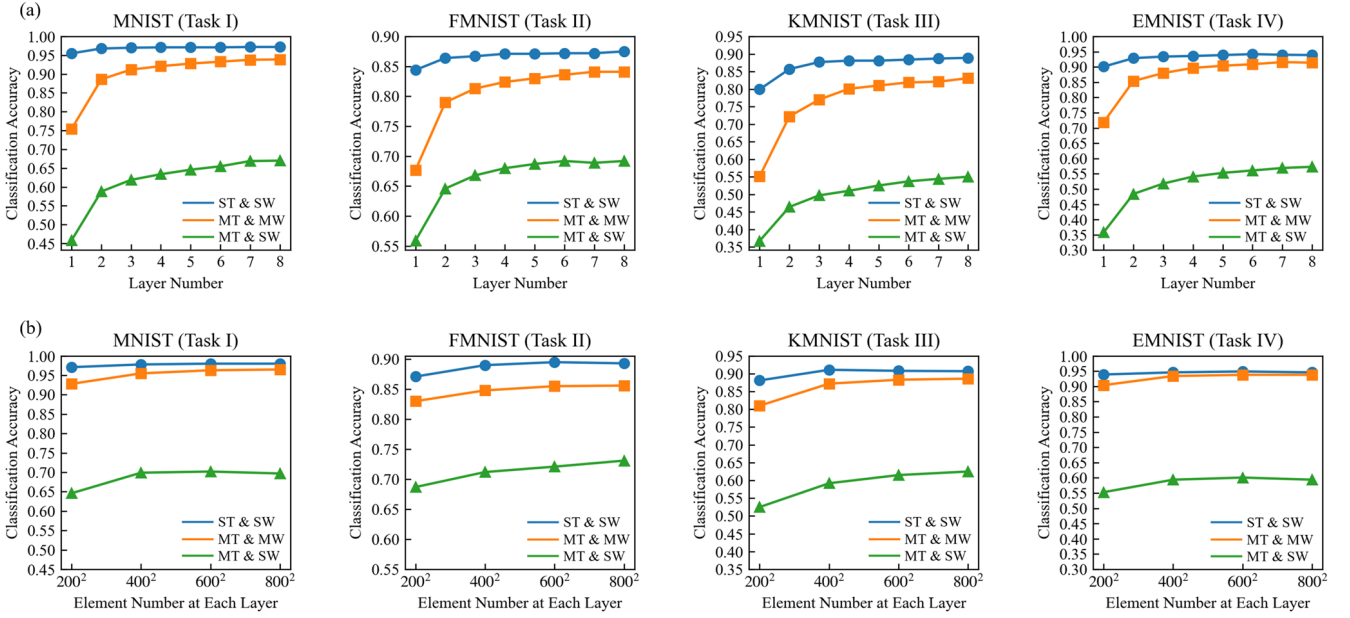

Fig. 4. The performance of four-wavelength D²NNs for four-task classification without wavelength selective filters. The classification accuracies of four-wavelength D²NNs in performing the four-task classification are significantly higher than the single-wavelength D²NNs. By increasing the layer number (a) and modulation element numbers (b) at each layer, the classification accuracies of four-wavelength D²NNs increase at each of four tasks and approach the classification accuracies by individually training four single-wavelength D²NNs to perform four tasks separately. ST & SW: single task using single wavelength; MT & SW: multi-task using single wavelength; MT & MW: multi-task using multiple wavelengths.

## 4 Discussion

There are different methods that have been widely validated for fabricating multi-wavelength diffractive optical elements (DOEs) to achieve the same phase modulation characteristic under different wavelength channels. [28-32]. The geometry structure of each element can be determined to make the optical path length of each modulation element have the same phase value for each wavelength. This can be achieved by adding an integral multiphase delay (e.g., 2π) to one wavelength until the other wavelength reaches the appropriate phase retardation [28, 29]. The overall physical height will be based on the actual accuracy requirements. The other method exploits the refractive index change of dispersive materials for different wavelengths. One can control the optical path length of each wavelength [30]. A multi-wavelength DOE can also be designed by combining several aligned DOEs, made of different materials, similar to the polarization-selective DOEs [31, 32]. Furthermore, the flexibility of wavefront manipulation in different physical dimensions, e.g., phase, amplitude, wavelength, and polarization, in the metasurfaces makes it possible to encode multiple wavelength channels. For example, a designed metasurface consisting of different types of nano-blocks with spatially varying rotation angles multiplexed in a subwavelength unit can make it resonant with different wavelengths [33].

The multi-wavelength D²NN is an all-optical computing processor that simultaneously performs multiple tasks with extremely low computing latency. The total computing time delay for each instance of multi-task input is the sum of wavefront propagation time from the input plane to the detector plane and the response time of the optoelectronic detector, which is independent of wavelength channel numbers and modulation element numbers at each layer. Taking the four-wavelength D²NN configured with five layers and 800×800 modulation element at each layer in Fig. 4 as an example, the total computing time for each instance of multi-task input is 1.23 ns by assuming a 30 GHz detection rate, which approximately performs the total diffractive optical computing operation numbers of 324 Million. Besides, compared with the spatial multiplexing of multiple individual D²NNs, the optical signals of different wavelength channels are independent of each other without any crosstalk during the multi-wavelength diffractive optical computing. Therefore, increasing the wavelength channels and modulation element numbers in a single monolithic system increases the computing throughput and facilitates more tasks.

Nowadays, artificial neural networks still cannot learn in a continuous manner like mammalian brains, which is a great hindrance to the development of general artificial intelligence. It is widely accepted that catastrophic forgetting is a necessary flaw in the connectionist model. Although machine learning algorithms, such as transfer learning, focus on storing knowledge gained while solving one problem and applying it to different but related problems, its essence is a mathematical process that ignores physical properties and can only achieve the serial processing of different tasks. Multi-wavelength $D^2$NN has the inherent advantages of parallel processing of multiple tasks with light-speed processing, low power consumption, and high throughput. By encoding different tasks into different wavelength channels, multi-wavelength $D^2$NN can significantly alleviate competition among different tasks and maintain high performance for each task. For each new task, a new wavelength channel can be added easily to implement the new task. The task expansion process is shown in Fig. 3(a), and the cost of the whole process is extremely low. Therefore, multi-wavelength $D^2$NN can take full advantage of photonic computing and is expected to support realizing a more general brain-inspired intelligence architecture in the future.

# 5  Conclusion

In this work, we have demonstrated the capability of multi-wavelength $D^2$NNs to achieve high-parallel classification and enable high-accuracy optical multi-task learning with the joint optimization training method. By encoding multi-tasks into multi-wavelength channels to exploit the wavelength dimension of the diffractive optical field, the proposed optical multi-task learning approach can realize different tasks in parallel at the speed of light. The optical multi-task function is implemented within a monolithic system and does not require the mechanical movement of diffractive modulation layers, significantly reducing the system's complexity. Analysis reveals that the proposed method can significantly alleviate the competition between multi-tasks and maintain the performance of each task. As the task number increases, the multi-wavelength $D^2$NNs show greater advantages in realizing optical multi-task learning. The proposed approach can be extended to other photonic neural network architectures by using the wavelength-division multiplexing technology to perform optical multi-task learning that simultaneously achieves the capability of high-parallel, high-accuracy, and high-generality.

# Author information

**Author contributions:** All the authors have accepted responsibility for the entire content of this submitted manuscript and approved submission.

**Conflict of interest statement:** The authors declare no conflicts of interest regarding this article.

**Research funding:** This work is supported by the National Key Research and Development Program of China (No. 2021ZD0109902), and the National Natural Science Foundation of China (No. 62275139).